\pdfoutput=1

\documentclass[11pt]{article}
\usepackage[table,xcdraw,dvipsnames]{xcolor}
\usepackage[]{style/acl2022/acl}

\usepackage{times}
\usepackage{latexsym}
\usepackage{graphicx}
\usepackage{tabularx}
\usepackage{makecell}
\usepackage{algorithm}
\usepackage{algpseudocode}
\usepackage{float}
\usepackage[T1]{fontenc}
\usepackage[utf8]{inputenc}
\usepackage{microtype}

\usepackage{CJKutf8}
\usepackage{amsmath}
\usepackage{amsmath}
\usepackage{booktabs}
\usepackage[textsize=footnotesize]{todonotes}   
\usepackage[english]{babel}
\newcommand{\chinese}[1]{\begin{CJK*}{UTF8}{gbsn}#1\end{CJK*}}
\newcommand{\abr}[1]{\textsc{#1}}
\usepackage{fullwidth}
\usepackage{cuted}

\title{Rare but Severe Neural Machine Translation Errors Induced by Minimal Deletion: An Empirical Study on Chinese and English}
\author{Ruikang Shi \and Alvin {Grissom II} \and Duc Minh Trinh \\
                Haverford College \\
                Haverford, PA, USA
                }

\begin{document}
\maketitle
\newcommand{\gl}[3]{%
\leavevmode\vtop{\hbox{#1}%
\hbox{#2\lower1.4ex\rlap{}}
\hbox{#3\lower1.4ex\rlap{}}
}}

\begin{abstract}
  We examine the inducement of rare but severe errors in
  English-Chinese and Chinese-English in-domain neural machine
  translation by minimal deletion of the source text with
  character-based models. By deleting a single character, we can
  induce severe translation errors. We categorize these errors and
  compare the results of deleting single characters and single
  words. We also examine the effect of training data size on the
  number and types of pathological cases induced by these minimal
  perturbations, finding significant variation. We find that deleting
  a word hurts overall translation score more than deleting a
  character, but certain errors are more likely to occur when deleting
  characters, with language direction also influencing the effect.

\end{abstract}

\section{Introduction}
Pathological machine translation (\abr{mt}) errors have been a problem
since the field's inception, and they have been analyzed and
categorized in the context of both statistical (\abr{smt}) and neural
machine translation (\abr{nmt}). Recent work examines pathologies in
\abr{nlp} models on classification problems: cases in which the models
make wildly inaccurate predictions, often confidently, when input
tokens are removed~\cite{feng-etal-2018-pathologies}. Identifying
these enriches our understanding of neural models and their points of
failure. \abr{mt} pathologies take the form of severe translation
errors, the worst being
hallucinations~\cite{lee2019hallucinations}. These rare errors are
difficult to study precisely because they are rare. In this paper, we
examine severe errors induced by minimal deletions by automatically
extracting translations with severe errors and manually categorizing
them.

Previous work taxonomizes \abr{smt}
errors~\cite{vilar-etal-2006-error} and analyzes their effects on
translation quality~\cite{federico2014assessing}. More recently,
\newcite{guerreiro2022looking} propose a taxonomy of \abr{mt}
pathologies, of which hallucinations are a
category.\footnote{\newcite{guerreiro2022looking} note that the term
  ``hallucination'' is overloaded and inconsistent; for this reason,
  we generally avoid the use of this term here.} They note the
shortcomings of current automatic detection methods, e.g., those based
on quality estimation and heuristics, and look for critical errors in
naturalistic settings.  They also propose \abr{DeHallucinator}, which
flags problematic translations and replaces them with re-ranking.

Other work on Chinese-English (Zh-En) \abr{smt} examines tense errors
caused by incorrectly translating \chinese{了}
(\textit{le})~\cite{liu-etal-2011-learning} and syntactic failures
caused by \chinese{的} (\textit{de}). More recent work uses input
perturbation to argue that \abr{nmt} models, including those based on
transformers~\cite{10.5555/3295222.3295349}, are brittle:
\citet{belinkov2018synthetic} examine the effect on \abr{nmt} systems
of several kinds of randomized perturbations by adding tokens, and
\newcite{niu2020evaluating} study subword regularization to increase
robustness to randomized perturbations. \newcite{raunak2021curious}
argue that memorized training examples are more likely to hallucinate,
and \newcite{voita2020analyzing} examine the contribution of source
and target tokens to errors. Also related, \newcite{sun2020adv}
suggest that \abr{bert} is less robust to misspellings than other
kinds of noise, which can occur naturalistically or through other
errors (e.g., encoding).

While we expect targeted adversarial examples---those explicitly
designed to cause a system to
fail~\cite{jia2017adversarial,ebrahimi2018adversarial}---to cause
serious errors, we focus on the ostensibly more benign case of
in-domain En$\leftrightarrow$Zh \abr{nmt} with minimal
deletions. Adding valid words introduces distractors with which the
\abr{mt} system must cope, while deleting words more often
\textit{removes} information without explicitly introducing lexical
distractors. Both are noise, but the latter is more naturally framed
as requiring recovery from missing information, while the former
introduces irrelevant and misleading information. At the character
level, this distinction is less clear, since both adding and removing
characters requires that the model translate despite unseen input
substrings---minimally corrupted inputs. Are minimal word or character
corruptions more harmful to a purely character-based \abr{nmt} model?
The answer is not obvious.

While most prior work examines western European languages, we examine
translation between Chinese and English, building upon work
identifying errors by observing change in
\abr{bleu}~\cite{BLEUarticle} after
perturbation~\cite{lee2019hallucinations}. But in contrast this prior
work, which adds tokens, we focus exclusively on single deletions to
examine \textbf{minimal conditions}---i.e., a missing character or
word, as in a typo or corruption---under which \textbf{severe errors}
are newly induced. For our purposes, a severe error leads to a
translation in which the original meaning is unrecoverable, but there
are others, as well~\cite{vilar-etal-2006-error}. For our purposes, we
use \abr{word changing} to cover these cases.

\section{Finding Candidates}

\begin{table*}[t!]
  \label{tab:error_types}
  \small
\centering
\begin{tabularx}{\textwidth}{l|>{\raggedright}X|p{3cm}}
\toprule
\textbf{Error Type} & \textbf{Example} & \textbf{Description}\tabularnewline
\midrule
\abr{word changing} &     
\textbf{Source:} Occupational health and occupational risks.\\
\textbf{Perturbed Source:} Occupational {\color{RoyalBlue}heath} and occupational risks\\
\textbf{Reference:}
      \gl{\chinese{职业}}{zhíyè}{occupational}
      \gl{\chinese{健康}}{jiànkāng}{health}
      \gl{\chinese{与}}{yǔ}{and}
      \gl{\chinese{职业}}{zhíyè}{occupational}
      \gl{\chinese{风险}}{fēngxiǎn}{risks}\\

  \textbf{Translation:}
  \gl{\chinese{职业}}{zhíyè}{occupational}
  \gl{\chinese{道德}}{dàodé}{\color{Mulberry}{ethics}}
  \gl{\chinese{和}}{hé}{with}
  \gl{\chinese{职业}}{zhíyè}{occupational}
  \gl{\chinese{危险}}{wéixiǎn}{dangers} & The model only mistranslates the perturbed word, leading to a simple error in which \textit{health} has been swapped with the unrelated word \textit{ethics} (which is also orthographically distant in the source text). 
 \tabularnewline
\midrule
\abr{inability} & 
\textbf{Source :} Christian Peace Action Groups.\\
    \textbf{Perturbed Source:} Christian {\color{RoyalBlue}PeaceAction} Groups.\\
    \textbf{Reference:}
    \gl{\chinese{基督教}}{jīdūjiào}{Christian}
    \gl{\chinese{和平}}{hépíng}{Peace}
    \gl{\chinese{行动}}{xíngdòng}{Action}
    \gl{\chinese{组织}}{zǔzhī}{Groups}\\
    \textbf{Translation:} {\color{blue}Christian} {\color{Mulberry}Peaction} {\color{blue}Groups} &
    Instead of outputting Chinese, the model copies English characters, including the nonsense word \textit{Peaction}.
    \tabularnewline
    \midrule
    \abr{missing parts} &
    \textbf{Source:} Residential institutions: services for children.\\
    \textbf{Perturbed Source:} {\color{RoyalBlue}esidential} institutions: services for children. \\
    \textbf{Reference:}
    \gl{\chinese{寄宿}}{jìsù}{Residential}
    \gl{\chinese{机构}}{jīgòu}{institutions}
    \gl{\chinese{：}}{:}{:}\gl{\chinese{为}}{wèi}{for}
    \gl{\chinese{儿童}}{értóng}{children}
    \gl{\chinese{提供}}{tígōng}{provide}
    \gl{\chinese{服务}}{fúwù}{services}\\
    \textbf{Translation:}
    \gl{\chinese{对}}{duì}{for}
    \gl{\chinese{儿童}}{er tóng}{children}
    \gl{\chinese{的}}{de}{`de'}
    \gl{\chinese{服务}}{fúwù}{services} &
    Only some of the text is translated. In this example, though the translation is interpretable, a substantial portion of the text is entirely untranslated.
    \tabularnewline
       \midrule
\abr{irrelevant} & 
\textbf{Source:} Maternal breastfeeding.\\
\textbf{Perturbed Source: } {\color{RoyalBlue}aternal} breastfeeding.\\
    \textbf{Reference:}
    \gl{\chinese{母乳}}{mǔrǔ}{maternal}
    \gl{\chinese{喂养}}{wèiyǎng}{breastfeeding}\\
    \textbf{Translation: }
    {
    \gl{\chinese{联合国}}{liánhéguó}{\abr{UN}}
    \gl{\chinese{维持}}{wéichí}{keep}
    \gl{\chinese{和平}}{hépíng}{peace}
    \gl{\chinese{行动}}{xíngdòng}{operation}
    \gl{\chinese{经费}}{jīngfèi}{funding}
    \gl{\chinese{的}}{de}{`de'}
    \gl{\chinese{筹措}}{chóucuò}{raise}} &
    This output is entirely hallucinated and has no apparent relationship to the input.
     \tabularnewline    
\bottomrule
\end{tabularx}
\caption{Examples and descriptions of triggers and error types found in low-scoring enumerations.
}
\label{tab:error_types}
\end{table*}
\label{sec:experiments}
 
We now describe the training of our \abr{nmt} model, method for
extracting severe error candidates (\textbf{enumerations}), and the
results of this extraction. For our extraction experiments, we begin
by examining character deletion before repeating the same experiments
with word deletion.  All experiments are done in both directions and
for two different training data sizes (1M and 10M
sentences), allowing us to observe the effect of training
data size, translation direction, and deletion type.

\subsection{Data and Models}
We train character-based En$\leftrightarrow$Zh models on the \abr{un}
Parallel Corpus 1.0~\cite{ziemski-etal-2016-united} of
sentence-aligned \abr{un} parliamentary documents.

We train two models in each direction with Sockeye
2~\cite{hieber-etal-2020-sockeye}---the first on the first 1M
sentences and the second on 10M---to observe the effect of training
data size on severe errors. We use the final 8,041 sentences as
validation and test data; the first 2,000 are test data.\footnote{We
  use a six-layer transformer with eight attention heads and a
  feed-forward network of 2,048 hidden units, trained on one 16GB
  Quadro P5000. Batch size is 256 and learning rate is .0002, reduced
  by a factor of .9 after 8 unimproving checkpoints. Training ceases
  when validation perplexity quiesces for 20 checkpoints of 4,000
  updates.  While \abr{bpe} has been shown to have higher \abr{bleu}
  on several datasets, this is not always the
  case~\cite{Cherry2018RevisitingCN}, and it can sometimes cause
  anomalies in translation
  itself~\cite{ataman2017linguistically,huck2017target}. We want to
  analyze the effect of deletion under simple conditions without this
  added complexity.}


\subsection{Identifying Error Candidates}
\begin{figure}[t!]
    \centering
    \includegraphics[width=\columnwidth]{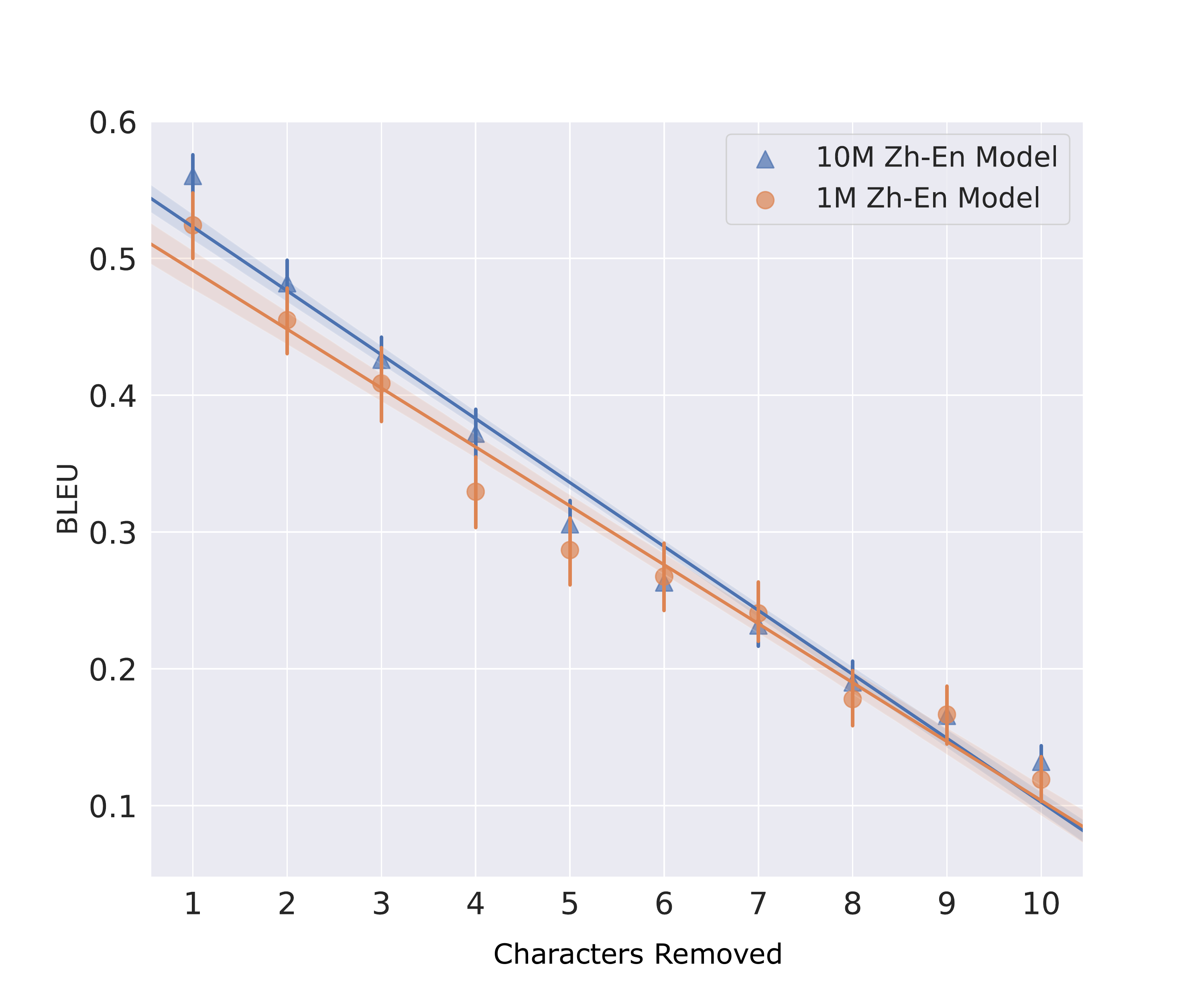}
    \caption{Zh-En \abr{bleu} as function of characters removed on valid
      sentences with 95\% confidence intervals. There is a linear
      relationship, with average \abr{bleu} converging as more tokens are
      removed.}
    \label{fig:zh-en-line}
\end{figure}

On translated test sentences, if sentence-level \abr{bleu} is above
0.5, the translation is considered \textbf{valid}.\footnote{We choose
  this because it is well above average \abr{bleu} for all models,
  resulting in an enumeration set with high average \abr{bleu}
  (Table~\ref{tab:all-results}), and few perturbed sentences reach
  this score (Figure~\ref{fig:zh-en-line}).} We translate valid
sentences with one token missing, exhaustively trying every
possible deleted token in every sentence. Perturbed sentences'
translations are called \textbf{enumerations}. If an enumeration's
sentence-level \abr{bleu} is less than .1, it is a candidate error, as
these precipitous drops are outliers in the linear decline in
\abr{bleu} as tokens are removed (Figure~\ref{fig:zh-en-line}). For a
more detailed specification of this process, see
Algorithm~{\ref{pseudocode}}, which is written for clarity rather than
efficiency.\footnote{Code is hosted \href{https://github.com/Shadoom7/HALLUCINATION_Research}{on GitHub}.}

\begin{algorithm}[hbt!]
\small
\caption{Algorithm for Finding Candidates. We describe the logic in
  three distinct steps for clarity, though there is obvious potential for optimization.}
\label{pseudocode}
\begin{algorithmic}[1]
\Function{Run}{\texttt{test\_sents}}
\State \texttt{valid} $\gets$ \texttt{find\_valid(test\_sents)}

\Comment{Find valid sentences with BLEU $\geq 0.5$.}
\State \texttt{generate\_enumerations(valid)}

\Comment{\parbox[t]{.65\columnwidth}{Iterate over every character in every valid sentence and try removing one character at a time. These perturbed sentences are enumerations.}}
\State \texttt{candidates} $\gets$ \texttt{find\_candidates(valid)}

\Comment{\parbox[t]{.65\columnwidth}{Keep the enumerations that have
    sentence-level BLEU $\leq 0.1$. We then manually identify those
    with actual severe errors and categorize them.}}  \EndFunction
\end{algorithmic}
\begin{algorithmic}[1] 
\Function{find\_valid}{\texttt{test\_sentences}}
    \For{each sentence \texttt{s} \textbf{in} \texttt{test\_sentences}}
        \State \texttt{s.bleu} $\gets$ \texttt{bleu(s)}
        \If{\texttt{s.bleu} $\geq 0.5$}
            \State \texttt{valid.add(s)}
        \EndIf
    \EndFor
    \State \textbf{return} valid
    \EndFunction
  \end{algorithmic}

\begin{algorithmic}[1]
\Function{generate\_enumerations}{\texttt{valid}}
    \For{each valid sentence \texttt{s} \textbf{in} \texttt{valid}}
        \For{each char index $i$ \textbf{in} \texttt{s}}
            \State \texttt{enum} $\gets$ \texttt{s.delete\_char(}$i$\texttt{)}
            \State \texttt{s.enums.add(enum)}
        \EndFor
        \EndFor
        \EndFunction
\end{algorithmic}

\begin{algorithmic}[1]
\Function{find\_candidates}{\texttt{valid}}
    \For{each sentence \texttt{s} \textbf{in} \texttt{valid}}
        \For{each \texttt{enum} \textbf{in} \texttt{s.enums}} 
            \State \texttt{new\_bleu $\gets$ bleu(enum)}
                \If{\texttt{new\_bleu} $\leq 0.1$}
                    \State \texttt{candidates.add(enum)}
                  \EndIf
           \EndFor
           \EndFor
           \State \textbf{return} candidates
           \EndFunction
\end{algorithmic}
\end{algorithm}


\section{Experiments and Results}

We now discuss our experiments and the results of our enumeration
extraction and the errors contained therein. All results are
summarized in Table~\ref{tab:all-results}, with results on the same
2,000 test sentences.

\subsection{Error Categorization}

We manually categorize errors into four types in our analysis:
\abr{word changing}, \abr{inability}, \abr{missing parts}, and
\abr{irrelevant}. Examples and descriptions are in
Table~\ref{tab:error_types}.\footnote{\abr{missing parts} differs from
  \abr{inability} in that what \textit{is} translated for \abr{missing
    parts} is correct. \abr{irrelevant} translations are readable but
  unrelated to the source, while \abr{inability} indicates a failure
  to generate readable output. It is possible in principle to have
  many types of errors in one bad translation, but we did not
  observe this.} What they have in common is that the original meaning
is unrecoverable, though simple \abr{word changing} is not considered
\textit{a priori} severe in our analysis.


     
    

\subsection{En-Zh 1M Training Sentence Results}
There are 96 candidate severe errors among 14,722 enumerations: ten
\abr{inability}, three \abr{irrelevant} and five \abr{missing
  parts}. The rest are \abr{word changing}. We have 18 errors
($.12\%$). \footnote{We also try removing sentences with English
  characters on the Chinese side, leaving 831,941 sentences on which
  to train. Translating these yields no \abr{inability} errors and
  leaves \abr{bleu} largely unchanged, suggesting that the
  untranslated named entities in the training data indeed cause
  \abr{inability}.  There are three \abr{missing parts} and two
  \abr{irrelevant} out of 63 potential hallucinations.  Test
  \abr{bleu} is largely unchanged, and valid \abr{bleu} decreases only
  slightly.}

One possible reason for these errors is that the model
has insufficient training data to generalize. We investigate by
training on ten times the data.

\subsection{En-Zh Model Trained on 10M Sentences}
We use the same corpus and architecture but use the first 10M instead
of 1M parallel sentences to train (En-Zh-10M). Validation perplexity
is nearly halved to 6.0 vs. the 1M model's 11.5 Likewise, \abr{bleu}
on the test data increases by .08 to .4
(Table~\ref{tab:all-results}), as expected. Unexpectedly, \abr{bleu}
on enumerations drops by .16 with more training data, much more than
the .11 drop with 1M training sentences, suggesting more training
data counterintuitively \textit{increases} sensitivity to minimal
character deletions, despite initial \abr{bleu} being
higher.



\begin{table*}[t]
  \centering
\resizebox{\textwidth}{!}{%
\begin{tabular}{lllllllllllll}
  \cellcolor[HTML]{C0C0C0}\textbf{Model} &
    \cellcolor[HTML]{91B8EF}\textbf{\abr{bleu}} &
  \cellcolor[HTML]{C0C0C0}\textbf{Deletion} &
  \cellcolor[HTML]{9BD39A}\textbf{Valid} &
  \cellcolor[HTML]{9BD39A}\textbf{\thead{\abr{bleu}\\(Valid)}} &
  \cellcolor[HTML]{DA9E83}\textbf{Enum.} &
  \cellcolor[HTML]{DA9E83}\textbf{\thead{\abr{bleu}\\(Enum.)}} &
  \cellcolor[HTML]{91B8EF}\textbf{$\Delta$\abr{bleu}} &
  \cellcolor[HTML]{C0C0C0}\textbf{In.} &
  \cellcolor[HTML]{C0C0C0}\textbf{\abr{mp}} &
  \cellcolor[HTML]{C0C0C0}\textbf{Irr.} &
  \cellcolor[HTML]{C0C0C0}\textbf{Total Errors} \\
  \cellcolor[HTML]{E2DDDD}En-Zh-1M &
  \cellcolor[HTML]{DAE8FC}.32 &
  \cellcolor[HTML]{E2DDDD}Char &
  \cellcolor[HTML]{C6E6C6}351 &
  \cellcolor[HTML]{C6E6C6}.77 &
  \cellcolor[HTML]{FFD6A6}14,722 &
  \cellcolor[HTML]{FFD6A6}.66 &
  \cellcolor[HTML]{DAE8FC} -.11 (-14.2\%) &
  \cellcolor[HTML]{E2DDDD}10 &
  \cellcolor[HTML]{E2DDDD}5 &
  \cellcolor[HTML]{E2DDDD}3 &
  \cellcolor[HTML]{E2DDDD}18 (0.12\%)\\
  \cellcolor[HTML]{EFEFEF}En-Zh-10M &
  \cellcolor[HTML]{ECF4FF}.40 &
  \cellcolor[HTML]{EFEFEF}Char &
  \cellcolor[HTML]{D8FAD8}506 &
  \cellcolor[HTML]{D8FAD8}.80 &
  \cellcolor[HTML]{FFE9CF}30,079 &
  \cellcolor[HTML]{FFE9CF}.64 &
  \cellcolor[HTML]{ECF4FF} -.16 (-20.0\%) &
  \cellcolor[HTML]{EFEFEF}33 &
  \cellcolor[HTML]{EFEFEF}0 &
  \cellcolor[HTML]{EFEFEF}0 &
  \cellcolor[HTML]{EFEFEF}33 (0.11\%) \\
  \cellcolor[HTML]{E2DDDD}Zh-En-1M &
  \cellcolor[HTML]{DAE8FC}.39 &
  \cellcolor[HTML]{E2DDDD}Char &
  \cellcolor[HTML]{C6E6C6}602 &
  \cellcolor[HTML]{C6E6C6}.73 &
  \cellcolor[HTML]{FFD6A6}11,093 &
  \cellcolor[HTML]{FFD6A6}.62 &
   \cellcolor[HTML]{DAE8FC}-.11 (-15.0\%) &
  \cellcolor[HTML]{E2DDDD}0 &
  \cellcolor[HTML]{E2DDDD}5 &
  \cellcolor[HTML]{E2DDDD}1 &
  \cellcolor[HTML]{E2DDDD}6 (0.05\%) \\
  \cellcolor[HTML]{EFEFEF}Zh-En-10M &
  \cellcolor[HTML]{ECF4FF}.42 &
  \cellcolor[HTML]{EFEFEF}Char &
  \cellcolor[HTML]{D8FAD8}714 &
  \cellcolor[HTML]{D8FAD8}.78 &
  \cellcolor[HTML]{FFE9CF}14,031 &
  \cellcolor[HTML]{FFE9CF}.67 &
   \cellcolor[HTML]{ECF4FF} -.11 (-14.1\%) &
  \cellcolor[HTML]{EFEFEF}0 &
  \cellcolor[HTML]{EFEFEF}1 &
  \cellcolor[HTML]{EFEFEF}0 &
  \cellcolor[HTML]{EFEFEF}1 (0.007\%) \\
  \cellcolor[HTML]{E2DDDD}En-Zh-1M &
  \cellcolor[HTML]{DAE8FC}.32 &
  \cellcolor[HTML]{E2DDDD}Word &
  \cellcolor[HTML]{C6E6C6}351 &
  \cellcolor[HTML]{C6E6C6}.77 &
  \cellcolor[HTML]{FFD6A6}2,521 &
  \cellcolor[HTML]{FFD6A6}.48 &
  \cellcolor[HTML]{DAE8FC} -.29 (-37.6\%) &
  \cellcolor[HTML]{E2DDDD}3 &
  \cellcolor[HTML]{E2DDDD}0 &
  \cellcolor[HTML]{E2DDDD}5 &
  \cellcolor[HTML]{E2DDDD}8 (0.32\%) \\
  \cellcolor[HTML]{EFEFEF}En-Zh-10M &
  \cellcolor[HTML]{ECF4FF}.40 &
  \cellcolor[HTML]{EFEFEF}Word &
  \cellcolor[HTML]{D8FAD8}506 &
  \cellcolor[HTML]{D8FAD8}.80 &
  \cellcolor[HTML]{FFE9CF}4,945 &
 \cellcolor[HTML]{FFE9CF}.54 &
  \cellcolor[HTML]{ECF4FF}-.26 (-32.5\%) &
  \cellcolor[HTML]{EFEFEF}7 &
  \cellcolor[HTML]{EFEFEF}0 &
  \cellcolor[HTML]{EFEFEF}2 &
  \cellcolor[HTML]{EFEFEF}9 (0.18\%) \\
  \cellcolor[HTML]{E2DDDD}Zh-En-1M &
  \cellcolor[HTML]{DAEAFC}.39 &
  \cellcolor[HTML]{E2DDDD}Word &
  \cellcolor[HTML]{C6E6C6}602 &
  \cellcolor[HTML]{C6E6C6}.74 &
  \cellcolor[HTML]{FFD6A6}6,666 &
  \cellcolor[HTML]{FFD6A6}.54 &
  \cellcolor[HTML]{DAEAFC} -.20 (-27.0\%) &
  \cellcolor[HTML]{E2DDDD}0 &
  \cellcolor[HTML]{E2DDDD}2 &
  \cellcolor[HTML]{E2DDDD}6 &
  \cellcolor[HTML]{E2DDDD}8 (0.12\%) \\  
  \cellcolor[HTML]{EFEFEF}Zh-En-10M &
  \cellcolor[HTML]{ECF4FF}.42 &
  \cellcolor[HTML]{EFEFEF}Word &
  \cellcolor[HTML]{D8FAD8}724 &
  \cellcolor[HTML]{D8FAD8}.78 &
  \cellcolor[HTML]{FFE9CF}8,461 &
  \cellcolor[HTML]{FFE9CF}.58 &
  \cellcolor[HTML]{ECF4FF} -.20 (-25.6\%) &
  \cellcolor[HTML]{EFEFEF}0 &
  \cellcolor[HTML]{EFEFEF}1 &
  \cellcolor[HTML]{EFEFEF}9 &
  \cellcolor[HTML]{EFEFEF}10 (0.11\%)
\end{tabular}%
}
\caption{Results of candidate extraction for minimal deletion,
  \abr{bleu} for each extracted set of sentences, and error statistics
  in models, broken down into \abr{inability} (\textbf{In.}),
  \abr{missing parts} (\textbf{\abr{mp}}), and \abr{irrelevant}
  (\textbf{Irr.}). \textbf{Valid} sentences with \abr{\abr{bleu}}
  $>0.5$ are extracted to create minimally perturbed
  \textbf{enumerations}; from these candidates, bad translations are
  extracted based on \abr{bleu} decline post-perturbation
  ($\Delta$\abr{bleu}). Despite character deletion introducing
  nonsense words into the input, word removal causes more of these
  severe errors. Surprisingly, despite Chinese characters containing
  more information, English deletion causes substantially
  higher decline in \abr{bleu}.}
\label{tab:all-results}
\end{table*}

There are 119 candidates among the 30,079 enumerations:
33 \abr{inability} and no \abr{missing parts} or \abr{irrelevant},
giving a $0.11\%$ probability of severe errors, approximately the same
as the 1M model ($0.12\%$).

The distribution of error types differs considerably when
training on more data: \abr{inability} errors triple.  We find that this is
due to untranslated words in the training data, all of which are named
entities.\footnote{By convention, sometimes named entities from
English are not translated into Chinese. \citet{ugawa2018neural} attempted
to improve \abr{nmt} with named entity tags to better handle compound
and ambiguous words, and other previous work showed that contamination
by another language~\cite{khayrallah2018} and copies of source
sentences in the target training data can degrade \abr{nmt}
performance.}  Since more training data contains more untranslated
named entities, \abr{inability} is more likely in models trained on
more data. We therefore train a model on the data where no English
appears in the references.

\subsection{Zh-En Experiments}
We examine Zh-En \abr{mt} under the same character deletion conditions
as En-Zh. Since Chinese characters contain more information than
English letters, we expect greater sensitivity to deletions on Zh-En,
but we do not find this (Table~\ref{tab:all-results}). Perturbing
En-Zh leads to consistently steeper declines in \abr{bleu}, as seen
in the valid vs. enumeration scores.





On the Zh-En model trained on 1M sentences, \abr{bleu} drops by .11,
from .73 for the 602 sentences to .62 for the enumerations, whereas
when trained on 10M sentences, we have .67 \abr{bleu} on enumerations,
which is higher than that of the smaller model. This is, notably, the
opposite of the En-Zh results, where more data decreased enumeration
\abr{bleu}. Both Zh-En experiments decrease by .11 \abr{bleu} on
enumerations, suggesting that the model with more training data is
similarly robust to this perturbation as the smaller model, unlike the
En-Zh case, in which the model trained on more data is more sensitive
to character perturbations.  As before, training models with more data
decreases Zh-En errors: on Zh-En model trained on 1M sentences, we
have 1 \abr{irrelevant} and 5 \abr{missing parts} (.05\%) errors,
while on when trained on10M sentences, we have 1 \abr{missing parts}
(.007\%). The remaining errors are \abr{word changing}.

There are no \abr{inability} errors in the two Zh-En experiments,
which accords with the results from En-Zh, suggesting that
\abr{inability} is due to the untranslated words in the training
data. Since there are no untranslated Chinese words on the English
side in the training data, we expect no \abr{inability} for a Zh-En
model.


\subsection{Minimal Word Deletion}
We now examine \textit{word} deletion as a basis of comparison. Does
the character \abr{nmt} model better handle the corrupted words caused by minimal
character deletion, or is it more robust to whole word deletion, which leaves coherent
words but removes more characters?\footnote{We use
  \abr{thulac}~\cite{sun2016thulac} \texttt{fast} for Chinese tokenization.}  We
find that, in all cases, deleting words leads to
\textit{substantially} lower \abr{bleu} than deleting characters, and
though still rare, confirmed severe error rates also increase.

For En-Zh trained on 1M sentences, for instance, \abr{bleu} for
enumerations drops to 0.48 in comparison to 0.66 when deleting
characters, and these stark differences in \abr{bleu}  persist.


On En-Zh trained on 1M sentences, we have 3 \abr{inability} and 5 \abr{irrelevant} ($.32\%$
severe errors). As expected, error rate increases considerably vs.
character removal ($.12\%$).


On En-Zh trained on 10M sentences, we have 7 \abr{inability} and 2 \abr{irrelevant}.
$.18\%$ of 4,945 enumerations are severe errors, also more likely than
with character deletion.



As with character deletion, increasing training size increases
\abr{inability} errors but decreases overall error probability. There
are no \abr{missing parts} errors when deleting words on En-Zh.

\subsection{Summary}
We see substantial variation in errors, depending on the kind of
deletion and translation direction, with \abr{inability} occurring
exclusively on En-Zh. We expect more \abr{bleu} decline on Zh-En,
since Chinese characters contain more semantic content and source
sentences are shorter, but we find the opposite of this with word
deletion.  We also find that while the models are more sensitive to
word deletion in terms of overall \abr{bleu}, this does not lead to
drastic increases in severe errors, suggesting that these severe
errors are unrelated to typical \abr{mt} errors, in line with
arguments that hallucinations should be considered separately from the
typical \abr{mt} errors~\cite{guerreiro2022looking}, due to the unique
patterns of heuristic-based methods when attempting to detect them.



\section{Conclusion and Future Work}
We examine the effect of minimal deletions on rare but severe \abr{mt}
errors on Chinese and English, using outlier changes in \abr{bleu}
after deletion to find candidates.

We find that the error rate for the model with a larger dataset is
always lower, suggesting more data can improve models' performance
against severe errors. Removing single words is more likely to cause
severe errors but less likely to cause \abr{missing parts} in our
models, despite character deletion introducing invalid words. On
En$\rightarrow$Zh, we observe none when removing words.  With the
important caveat that these errors are already rare, limiting the
conclusions we can make, this may suggest that Zh$\leftrightarrow$En
models are better able to recover when characters are missing, even
if the substrings themselves have never been observed, despite not
having been trained with such noise. This is not obvious for a
character-based model. Nor is it obvious that Zh$\rightarrow$En models
will be more robust to perturbations than En$\rightarrow$Zh, but this
is what we find, especially for words, perhaps because English words
are simply longer. Furthermore, that $\Delta$\abr{bleu} is not
predictive of significantly more severe errors suggests that these
errors are a different phenomenon from typical \abr{mt} shortcomings.

Further research is needed to determine the effect various variables
on robustness with targeted probes; future work can also determine how
findings generalize across more language pairs (potentially
typologies), tokenization schemes, and architectures. Training models
with missing source \textit{words} may increase robustness. For
detection, unusually large disparities in length between source and
target could signal \abr{inability} or \abr{missing parts} errors, and
vocabulary or semantic distance checks could flag bad translations
(e.g., \abr{word changing}, \abr{inability}).  It would also be
instructive to examine the extent to which \abr{nmt} robustness to
noise mirrors that of humans.



\section*{Acknowledgements}
We thank John Dougherty and Sorelle Friedler for their helpful
feedback.

\bibliographystyle{../style/acl2022/acl_natbib}
\bibliography{bib/2022-hallucinations-coling}



\end{document}